\documentclass[manuscript]{acmart}

\usepackage{color, colortbl}
\usepackage{amsmath}
\usepackage{array}
\usepackage{keystroke}
\usepackage{makecell}
\usepackage{pifont}
\usepackage{graphicx}
\usepackage{booktabs} 
\usepackage{array}
\usepackage{multirow}
\usepackage[export]{adjustbox}
\usepackage{pifont}
\usepackage{balance}
\usepackage{subfigure} 
\usepackage{tabularx}
\usepackage{menukeys} 
\usepackage{enumitem}
\usepackage{caption}
\usepackage{fontawesome}

\definecolor{materialteal}{HTML}{009688}
\definecolor{materialpurple}{HTML}{9C27B0}
\definecolor{materialindigo}{HTML}{3F51B5}
\definecolor{materialblue}{HTML}{2196F3}

\AtBeginDocument{%
  \providecommand\BibTeX{{%
    \normalfont B\kern-0.5em{\scshape i\kern-0.25em b}\kern-0.8em\TeX}}}

\def \debug{}
\ifx \debug \undefined
    \newcommand{\fx}[1]{{}}
\else
    \newcommand{\fx}[1]{{\textcolor{red}{#1}}}

\newcommand{\xdownarrow}[1]{%
  {\left\downarrow\vbox to #1{}\right.\kern-\nulldelimiterspace}
}

\definecolor{materialcyan}{HTML}{00BCD4}
\definecolor{materialteal}{HTML}{009688}
\definecolor{materialgreen}{HTML}{4CAF50}
\definecolor{materiallime}{HTML}{CDDC39}
\definecolor{materialamber}{HTML}{FFC107}
\definecolor{materialbrown}{HTML}{795548}
\definecolor{materialred}{HTML}{FF4436}
\definecolor{materialorange}{HTML}{FF5722}

\def \billahdebug{}

\ifx \billahdebug \undefined
\newcommand{\fixmesb}[1]{{}}
\newcommand{\fixmeth}[1]{{}}
\else
\newcommand{\fixmesb}[1]{{\bf\textcolor{red}{ [ sB FIXME: #1 ]}}}
\newcommand{\fixmeth}[1]{{\bf\textcolor{blue}{ [ tH FIXME: #1 ]}}}

\fi

\newcolumntype{L}[1]{>{\raggedright\let\newline\\\arraybackslash\hspace{0pt}}m{#1}}
\newcolumntype{C}[1]{>{\centering\let\newline\\\arraybackslash\hspace{0pt}}m{#1}}
\newcolumntype{R}[1]{>{\raggedleft\let\newline\\\arraybackslash\hspace{0pt}}m{#1}}
\newcolumntype{B}{>{\bfseries}c}

\copyrightyear{2024}
\acmYear{2024}
\setcopyright{rightsretained}
\acmConference[ASSETS '24]{The 26th International ACM SIGACCESS Conference on Computers and Accessibility}{October 27--30, 2024}{St. John's, NL, Canada}
\acmBooktitle{The 26th International ACM SIGACCESS Conference on Computers and Accessibility (ASSETS '24), October 27--30, 2024, St. John's, NL, Canada}
\acmDOI{10.1145/3663548.3688538}
\acmISBN{979-8-4007-0677-6/24/10}

\begin{document}


\title[Identifying Crucial Objects]{Identifying Crucial Objects in Blind and Low-Vision Individuals' Navigation}

\author[MT Islam]{Md Touhidul Islam}
\affiliation{%
  \institution{Pennsylvania State University}  
  \city{University Park}
  \state{PA}
  \country{USA}
}
\email{touhid@psu.edu}

\author[I Kabir]{Imran Kabir}
\affiliation{%
  \institution{Pennsylvania State University}  
  \city{University Park}
  \state{PA}
  \country{USA}
}
\email{ibk5106@psu.edu}

\author[EA Pearce]{Elena Ariel Pearce}
\affiliation{%
  \institution{Drake University}  
  \city{Des Moines}
  \state{IA}
  \country{USA}
}
\email{elena.pearce@drake.edu}

\author[MA Reza]{Md Alimoor Reza}
\affiliation{%
  \institution{Drake University}  
  \city{Des Moines}
  \state{IA}
  \country{USA}
}
\email{md.reza@drake.edu}

\author[SM Billah]{Syed Masum Billah}
\affiliation{%
  \institution{Pennsylvania State University}  
  \city{University Park}
  \state{PA}
  \country{USA}
}
\email{sbillah@psu.edu}

\begin{abstract}
This paper presents a curated list of 90 objects essential for the navigation of blind and low-vision (BLV) individuals, encompassing road, sidewalk, and indoor environments. We develop the initial list by analyzing 21 publicly available videos featuring BLV individuals navigating various settings. Then, we refine the list through feedback from a focus group study involving blind, low-vision, and sighted companions of BLV individuals. A subsequent analysis reveals that most contemporary datasets used to train recent computer vision models contain only a small subset of the objects in our proposed list. Furthermore, we provide detailed object labeling for these 90 objects across 31 video segments derived from the original 21 videos. Finally, we make the object list, the 21 videos, and object labeling in the 31 video segments publicly available. This paper aims to fill the existing gap and foster the development of more inclusive and effective navigation aids for the BLV community.
\end{abstract}





\maketitle
\section{Motivation}
Navigating urban environments presents significant challenges for blind and low-vision (BLV) individuals. 
While advancements in computer vision offer potential solutions through real-time object detection systems, existing datasets for training the underlying models in these systems often lack accessibility-specific annotations. 
Most contemporary datasets, such as ImageNet~\cite{deng2009imagenet} and MS-COCO~\cite{lin2014microsoft}, though extensive, do not include critical objects and features that are essential for BLV navigation, such as curb cuts, sidewalk conditions, and specific indoor landmarks. This gap hinders the development of robust navigation aids tailored to the needs of BLV individuals. 

To address this issue, we curate a list of 90 objects crucial for BLV navigation by analyzing 21 publicly available videos featuring BLV individuals in various settings. 
This list is later refined through feedback from focus groups involving BLV individuals and their sighted companions. 
Our analysis indicates that contemporary datasets cover only a small subset of our proposed objects. 
We also provide object labeling for the 90 objects across 31 video segments derived from the original videos. 
By making our object list, the 21 videos, and the labeled video segments publicly available~\cite{islam2024dataset}\footnote{\href{https://github.com/Shohan29531/BLV-Road-Nav-Accessibility}{https://github.com/Shohan29531/BLV-Road-Nav-Accessibility}}, we aim to bridge this gap and support the development of more inclusive navigation aids for the BLV community. 

\section{Background and Related Work}
Large datasets contain thousands of images annotated with object/class names, bounding rectangles around objects, textual descriptions, and per-pixel semantic labels. 
As AI tasks evolve—from object classification and detection to scene understanding, VQA, and semantic segmentation—the effort in annotating data increases, leading to smaller datasets.
ImageNet~\cite{deng2009imagenet} is a classic image classification dataset with over 14M images and 1000 classes. In contrast, MS-COCO~\cite{lin2014microsoft}, popular for scene understanding, has 328K images and 91 classes. 
Visual Question Answering (VQA) datasets are an exception; for example, VQA v2.0~\cite{balanced_vqa_v2} has 200K images, 614K questions, and 10 answers per question. 
Semantic segmentation datasets are much smaller due to the laborious nature of pixel-wise annotation~\cite{mapiliary_iccv17, kitti15_ijcv18, cityscapes_cvpr16}.


Most large-scale, publicly available datasets are not accessibility-aware, with a few exceptions (e.g.,\cite{gurari2018vizwiz, saha2019project, theodorou2021disability}). 
They lack annotations crucial for people with disabilities. 
For instance, snow on sidewalks, areas near curb cuts, and fire hydrants are essential safety information for BLV individuals but such details are missing in mainstream datasets. 
Theodorou et al. reported that models trained on generic datasets fail to detect objects in pictures taken by blind users, as these images tend to be blurrier and may show out-of-frame objects\cite{theodorou2021disability}. 
The KITTI-15~\cite{kitti15_ijcv18} dataset, used for evaluating AI in autonomous driving, also lacks accessibility-aware annotations like curb cuts, ramps, and crosswalks.

The lack of accessibility awareness in large-scale public datasets is consequential but unsurprising. Most annotators are sighted crowd workers who often lack a nuanced understanding of disability~\cite{tigwell2021nuanced} and may ignore individuals with disabilities as "outliers" or label them generically (e.g., "a person with a disability")~\cite{linton1998claiming, davis2016theDS, morris2020ai}.
Incorporating accessibility awareness in datasets is challenging. Prior attempts~\cite{theodorou2021disability, park2021designing} revealed several challenges: engaging with target communities, providing accessible data collection tools, ensuring data quality, and balancing data structure with demands on collectors. Consequently, accessibility-focused datasets like VizWiz~\cite{gurari2018vizwiz} have significantly fewer annotations (e.g., 30K images in VizWiz vs. over 200K in common VQA datasets~\cite{balanced_vqa_v2}).

A possible workaround is to extract data from publicly available video-sharing platforms like YouTube and Vimeo. Previous studies have analyzed YouTube videos to understand people with vision impairments~\cite{xie2022youtube, seo2017exploring, seo2021understanding, seo2018understanding}. Following this approach, we collected 21 videos from YouTube and Vimeo featuring BLV individuals.

\section{Identification of Objects with Accessibility Impact}
\label{sec:taxonomy_dataset}

\quad\textbf{Video Collection.}
We collected free, publicly available videos from YouTube and Vimeo using a systematic keyword-based search. 
Some keywords used in the search were ``blind'', ``vision impairment'', and ``visually impaired'', along with the terms  ``white cane'', ``navigating'', ``orientation and mobility'', and ``training''. 
We briefly reviewed the videos from search results and selected videos relevant to blind navigation. 
In total, we collected 21 suitable videos, 16 from YouTube and 5 from Vimeo (Table~\ref{table:dataset}).

\textbf{Ethical Considerations.}
As per the Common Rule of the federal regulations on human subjects research protections (45 CFR 46.104(d)(4)), the collection of existing data is exempt from IRB review when the sources are publicly available. Nonetheless, we contacted the video uploaders on YouTube and Vimeo to give them credit and to solicit their permission.

\textbf{Identification of Objects with Accessibility Impact.}
Two researchers watched all videos and noted when relevant objects first appeared in the video in a Google Sheet. 
An object is deemed relevant 
(a) if it is on the way of the blind individual featured in the video and affects their course (e.g., they changed direction, they collided with the object); 
(b) if it provides meaningful feedback (e.g., produces different sounds or tactile feedback); 
(c) if it affects them physically (e.g., they tripped); 
(d) if it could be a hindrance in the future (e.g., the electrical box on a powered gate that sticks out into the sidewalk);
(e) if it could be mislabeled as something else (e.g., the escalator could be mislabeled as stairs, which could lead to someone being hurt from not expecting them to be moving); and
(f) if it could be thought of or labeled in the wrong context (e.g., a parked car on the sidewalk could lead a person to believe they were in the street).

This process generated around 80 object categories related to the accessibility of sidewalks and non-visual navigation.

\vspace{-10pt}
\section{Revising Objects with Accessibility Impact: A Focus Group Study}
\label{sec:tax}

We conducted a focus group study with Six participants to revise our initial list of 80 objects and determine whether any objects were missing or redundant.
This section presents the study's methodology, results, and implications for design.

\vspace{-15pt}
\subsection{Procedure}
\quad\textbf{Participants.}
Among the six study participants: 
Two were blind, two were low-vision, and two were sighted individuals.
The two sighted participants were experts in non-visual navigation because of their professions and proximity to blind individuals.
Participants were chosen based on their familiarity with AI-based applications, such as ``Seeing AI''~\cite{seeingAI}, ``Be My Eyes''~\cite{bemyeyes}, and ``Oko''~\cite{oko_app}. 
Among the six, two were males. 
The average age was 49.17 (SD 7.45).

Two researchers conducted the study; 
while one researcher asked predefined questions on possibly important objects (from the initial list) for accessible navigation, 
the other researcher took notes and identified new objects of interest.
The session lasted around an hour.
%

\textbf{Prompts.}
We established common ground by describing an imaginary AI app similar to ``Seeing AI''~\cite{seeingAI} and ``Aira''~\cite{Aira} that can detect and read out surrounding objects. 
We generated the following questions/prompts:
\textbf{\textit{Prompt 1}}: \emph{Let us imagine we have an AI app that can read out a scene in front of you. What objects would you want the AI to describe proactively?}
\textbf{\textit{Prompt 2}}: \emph{We have compiled a list of around 80 objects that we felt would be important in navigation for blind individuals. Would you please mention whether an object is relevant or not when we read it out?}
\textbf{\textit{Prompt 3}}: \emph{Please add more objects that are not on our list, but you feel would be important for the scenario (i.e., non-visual navigation). Please also provide a rationale for your choice.} 
    
\vspace{-10pt}
\subsection{Feedback on the Initial List}
\label{subsec:findings}

Below, we summarize study participants' object-specific opinions and the relative priority of the objects in our preliminary list. 

\textbf{Sidewalk Objects.}
Participants unanimously agreed on the importance of detecting objects that share the sidewalk, such as \textit{bicycles}, \textit{wheelchairs}, \textit{pets}, and \textit{water hoses}. 
\textit{Bicycles} and \textit{pets} were highlighted as top priorities. 
While one participant noted that a loose \textit{water hose} is a tripping hazard, two others felt their cane could detect it, preferring the AI to focus on other objects.

\textbf{Sidewalk Obstructions.}
Participants marked all sidewalk obstructions as important, with \textit{growing tree branches} and \textit{barrier posts} receiving the highest priority.


\textbf{Traffic Signals and Signs.}
Participants unanimously agreed on the importance of detecting \textit{traffic signals} and \textit{signs}, except for speed limit signs, which they deemed unnecessary for blind individuals. 
They emphasized the need for accurate detection of \textit{pedestrian crossing signals}, especially with aural warnings, citing apps like "Oko"~\cite{oko_app}. 
However, they expressed concerns about over-reliance on AI tools, with one participant highlighting the importance of including legal disclaimers, referencing the lengthy terms of the Oko app.

\textbf{Indoor Objects.}
Participants emphasized the need for accurate indoor object detection, particularly for \textit{chairs}, \textit{tables}, \textit{elevators}, and \textit{moving walks}, which are crucial in venues like restaurants, theaters, and public transport. 
While \textit{escalators} and \textit{stairs} were considered important, they were considered less critical due to their prominence and ease of detection.

\vspace{-5pt}
\subsection{Revised Taxonomy and Design Implications}
\label{subsec:design_implications}

This section presents the revised object taxonomy and discusses design implications for an AI application supporting navigation for blind and low-vision individuals.

\textbf{Newly Identified Objects.}
Throughout our study, participants frequently named items they considered important for blind and low-vision navigation. 
Many items matched our list, often with slight name variations. 
Discussions also revealed new objects, which we added to our revised taxonomy, such as \textit{doorways}, \textit{black ice}, \textit{cobblestone pavements}, and \textit{moving walks}. 
Participants found some items redundant, which we removed. We finalized a list of 90 objects, categorized in Table~\ref{table:taxonomy}.

\begin{table*}[!ht]
    \centering
    \resizebox{!}{3cm}{
\begin{tabular}{B | L{7cm} | C{11.8cm} } 

  \hline
   \textbf{\textsc{Group}} & 
   \textbf{\textsc{Parent Concept}} & 
   \textbf{\textsc{Accessibility-related objects}}
   \\ 
  \hline
\rowcolor{gray!10} 
1 & Attributes of a sidewalk and driveway & Accent Paving, Driveway (flat), Puddle, Raised Entryway, Sidewalk, Sidewalk Pits, Sloped Driveway, Tactile Paving, Brick Paving, Cobblestone Paving, Unpaved Sidewalk, Wet Surface\\ \hline
2 & Obstructions likely to be detected by a white cane & Fire hydrant, Gutter, Vegetation,
Tree, Brick Wall, Fence, Trash Bins, Lamp Post, Pole, Mailbox\\ \hline
\rowcolor{gray!10} 
3 & Obstructions less likely to be detected by a white cane & Closed Sidewalk, Barrier Post, Barrier Stump, Foldout Sign, Bench\\ \hline
4 & Objects that are too late to be detected by a white cane & Train Tracks, Train Platform\\ \hline
\rowcolor{gray!10} 
5 & Objects that pick you before you pick them & Overhanging Tree Branches\\ \hline
6 & Objects that provide navigational guidance & Retaining Wall, Railing, Wall, Curved Railing\\ \hline
\rowcolor{gray!10} 
7 & Objects not supposed to be on the sidewalk & Hose, Maintenance Vehicle, Trash on Roads, Snow, Water Leakage, Yard Waste, Water Pipes\\ \hline
8 & Moving objects sharing the sidewalk & Person, Bicycle, Wheelchair, Person with a Disability, White Cane, Dog, Guide Dog, Street Vendor \\ \hline
\rowcolor{gray!10} 
9 & Intersection & Pedestrian Crossing, Slopped Curb, Intersection, Crosswalk, Curb, Bridge, Uncontrolled Crossing\\ \hline
10 & Objects on the road shoulder & Road Shoulder, Roadside Parking, Parallel Parking Spot, Paratransit Vehicle\\ \hline
\rowcolor{gray!10} 
11 & Objects on the road & Road, Unpaved Road, Bus, Car, Motorcycle, Road Divider
\\ \hline
12 & Traffic signals and street signs & Traffic Signals, Stop Sign, Sign, Sign Post, Push Button, "Use the Other Door" Sign, Toilet Sign  \\ \hline
\rowcolor{gray!10} 
13 & Objects related to building exits and entrances & Gate, Flush Door, Doorway \\ \hline
14 & Indoor objects & Counter, Elevator, Escalator, Fountain, Stairs, Uneven Stairs, Table, Building, Moving Walk, Pillar, Chair\\ \hline
\rowcolor{gray!10} 
15 & Objects related to public transit & Bus Stop, Turnstile\\ \hline
\end{tabular}
}
    \caption{Key accessibility-related objects, classified into different groups.}
    \label{table:taxonomy}
\end{table*}

\vspace{-15pt}
\textbf{AI Tools Are Not a Replacement of Physical Assistance Devices.}
Our study participants emphasized that AI tools should not replace common physical assistance devices such as white canes.
One said:
\emph{``I have my concerns about using a phone [all the time] for lack of reliability. Having a physical object such as a white cane to guide is always preferred.''}
Our participants also reported that while they would appreciate the detection of some objects via AI beforehand, 
they want their cane, which they trust more, to take care of most detection tasks.

\textbf{Information Priority.}
Our participants agreed that navigation-related objects should be prioritized. 
For instance, they noted that elevators are more critical than escalators in complex indoor setups due to difficulty locating elevators. 
They suggested ignoring trash smaller than four inches and small pits that do not disrupt walking. 
Wet surface warnings are redundant if it is already raining. The presence of a driveway is crucial, but whether it is flat or sloped is less critical. 
Vehicles on the road should not be detected. 
And if an object is within reach of a white cane, AI assistance should remain silent.

\textbf{Configurable Information Presentation.}
Participants desire the ability to choose which objects they are alerted to and for the AI tool to learn from their usage habits, such as turning off specific object detection frequently opted out by the user. 
They also want the tool to adjust its detection and recommendations based on their mode of travel (e.g., with a guide dog, with a white cane, or in specific weather conditions).

One participant wanted the tool to provide weather-appropriate dressing advice, including heavy jackets or raincoats recommendations. 
Others believed the tool should issue cautions if the weather is ideal for icing and emphasized the importance of detecting ice beforehand, though they acknowledged the difficulty for AI, given the challenge for humans.
One participant suggested the tool could use recent weather data and temperature warnings near the icing point to predict which roads might have ice.

\textbf{Objects that Pick You Before the Cane Picks Them.}
Participants reported specific objects creating significant navigational challenges for blind and low-vision individuals. Tree branches extending over sidewalks pose a significant threat, often existing at head level or higher and undetectable by canes, leading to frequent injuries. 
Another challenging object is a train track. One participant noted that detecting a train track with a cane when a train is approaching is already too late.

\section{Analyzing the Object List: Coverage in Prominent Datasets}

\begin{figure*}[!ht]
    \centering
    \includegraphics[width=1.0\linewidth]{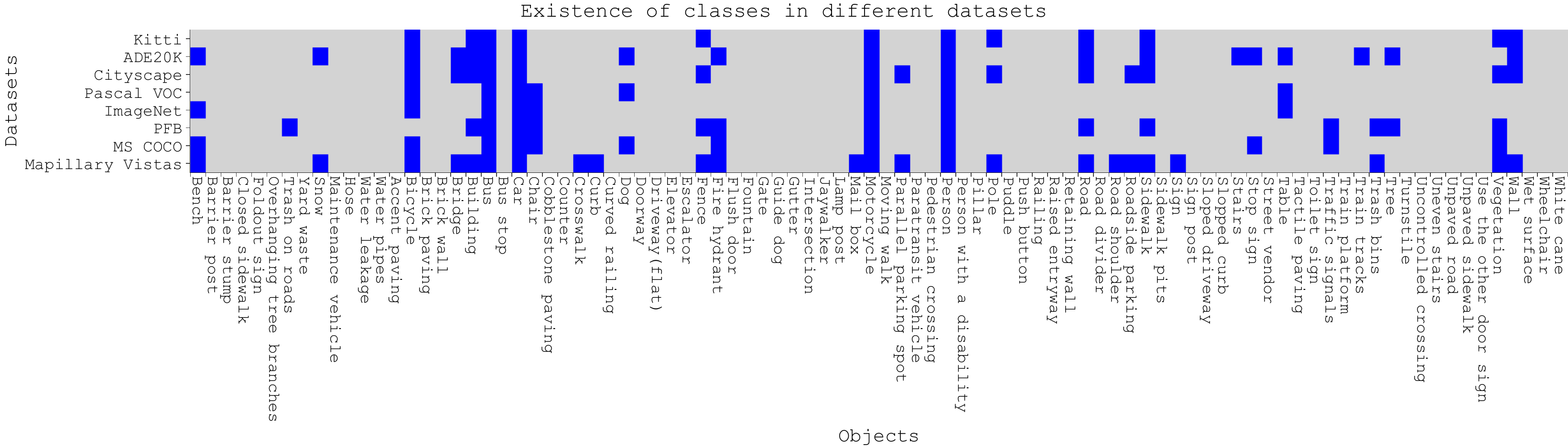}
    \caption{A heatmap representing the existence of different objects of our list in prominent datasets. \colorbox[HTML]{0000FF}{\phantom{x}} in a cell 
    means the corresponding object exists in the corresponding dataset. 
    In contrast, \colorbox[HTML]{D3D3D3}{\phantom{x}} means the object does not exist in the corresponding dataset. }
    \label{fig:obj_existance}
\end{figure*}

An AI model's effectiveness in assisting BLV individuals' navigation hinges on accurately detecting objects from our proposed list. 
However, object recognizers are limited to items in their training datasets, such as ImageNet~\cite{deng2009imagenet}, MS COCO~\cite{lin2014microsoft}, Mapillary Vistas~\cite{neuhold2017mapillary}, Kitti~\cite{geiger2012we}, Cityscapes~\cite{cordts2016cityscapes}, Pascal VOC~\cite{everingham2010pascal}, PFB~\cite{pfb_ICCV2017}, and ADE20K~\cite{zhou2017scene, zhou2019semantic}. 
Among these, Mapillary Vistas is the most advanced, offering detailed annotations for 66 outdoor object categories~\cite{neuhold2017mapillary}, 
but it and other datasets lack coverage for many objects on our list. 
Figure~\ref{fig:obj_existance} highlights this gap in these datasets.

Our study findings indicate that objects in groups such as "\textit{Obstructions less likely to be detected by a white cane (group 3)}", "\textit{Objects that pick you before you pick them (group 5)}", and "\textit{Objects not supposed to be on the sidewalk (group 7)}" are the most significant. Failure to detect these objects can lead to accidents or injuries for blind individuals.

The first nine objects in Figure~\ref{fig:obj_existance} (from \textit{Bench} to \textit{Water Pipes}) are the most significant objects that belong to the aforementioned groups 3, 5, and 7. 
Only three of these objects exist in some datasets, such as Mapillary Vistas, ADE20K, ImageNet, and PFB---highlighting their limitations in this aspect.

\section{Object Labeling}
We picked 31 video segments from 21 videos based on their diversity of content, object density, and key transitions and manually created their object labeling.
The authors visually inspected and annotated keyframes by labeling objects' presence (1) or absence (0) in each frame, using a differential approach that took less than 60 seconds per frame after the first keyframe. 
We make these object-labeling, the 21 videos, and the 90 objects publicly available~\cite{islam2024dataset}.
Appendix~\ref{subsec:object-labeling} contains more details regarding the labeling process.
Preliminary evaluations of seven well-known models using our object labeling are presented in Appendix~\ref{sec:assessing_ai_models}.

\bibliographystyle{ACM-Reference-Format}
\bibliography{Bibliography, Bibliography2, Bibliography3}


\begin{thebibliography}{41}


\ifx \showCODEN    \undefined \def \showCODEN     #1{\unskip}     \fi
\ifx \showDOI      \undefined \def \showDOI       #1{#1}\fi
\ifx \showISBNx    \undefined \def \showISBNx     #1{\unskip}     \fi
\ifx \showISBNxiii \undefined \def \showISBNxiii  #1{\unskip}     \fi
\ifx \showISSN     \undefined \def \showISSN      #1{\unskip}     \fi
\ifx \showLCCN     \undefined \def \showLCCN      #1{\unskip}     \fi
\ifx \shownote     \undefined \def \shownote      #1{#1}          \fi
\ifx \showarticletitle \undefined \def \showarticletitle #1{#1}   \fi
\ifx \showURL      \undefined \def \showURL       {\relax}        \fi
\providecommand\bibfield[2]{#2}
\providecommand\bibinfo[2]{#2}
\providecommand\natexlab[1]{#1}
\providecommand\showeprint[2][]{arXiv:#2}

\bibitem[see({[n.\,d.]})]%
        {seeingAI}
 \bibinfo{year}{[n.\,d.]}\natexlab{}.
\newblock \bibinfo{title}{Seeing AI}.
\newblock
\newblock
\urldef\tempurl%
\url{https://www.microsoft.com/en-us/seeing-ai/}
\showURL{%
\tempurl}


\bibitem[bem(2015)]%
        {bemyeyes}
 \bibinfo{year}{2015}\natexlab{}.
\newblock \bibinfo{title}{Be My Eyes: Bringing sight to blind and low vision
  people}.
\newblock
\newblock
\urldef\tempurl%
\url{https://www.bemyeyes.com/}
\showURL{%
\tempurl}


\bibitem[Abu~Alhaija et~al\mbox{.}(2018)]%
        {kitti15_ijcv18}
\bibfield{author}{\bibinfo{person}{Hassan Abu~Alhaija},
  \bibinfo{person}{Siva~Karthik Mustikovela}, \bibinfo{person}{Lars Mescheder},
  \bibinfo{person}{Andreas Geiger}, {and} \bibinfo{person}{Carsten Rother}.}
  \bibinfo{year}{2018}\natexlab{}.
\newblock \showarticletitle{Augmented reality meets computer vision: Efficient
  data generation for urban driving scenes}.
\newblock \bibinfo{journal}{\emph{International Journal of Computer Vision}}
  \bibinfo{volume}{126}, \bibinfo{number}{9} (\bibinfo{year}{2018}),
  \bibinfo{pages}{961--972}.
\newblock


\bibitem[Aira(2018)]%
        {Aira}
\bibfield{author}{\bibinfo{person}{Aira}.} \bibinfo{year}{2018}\natexlab{}.
\newblock \bibinfo{title}{Aira}.
\newblock \bibinfo{howpublished}{\url{https://aira.io/}}.
\newblock
\urldef\tempurl%
\url{https://aira.io}
\showURL{%
Retrieved May 23, 2020 from \tempurl}


\bibitem[Antol et~al\mbox{.}(2015)]%
        {antol2015vqa}
\bibfield{author}{\bibinfo{person}{Stanislaw Antol}, \bibinfo{person}{Aishwarya
  Agrawal}, \bibinfo{person}{Jiasen Lu}, \bibinfo{person}{Margaret Mitchell},
  \bibinfo{person}{Dhruv Batra}, \bibinfo{person}{C~Lawrence Zitnick}, {and}
  \bibinfo{person}{Devi Parikh}.} \bibinfo{year}{2015}\natexlab{}.
\newblock \showarticletitle{{VQA}: {Visual} {Question} {Answering}}. In
  \bibinfo{booktitle}{\emph{Proceedings of the IEEE International Conference on
  computer vision}}.
\newblock


\bibitem[Cordts et~al\mbox{.}(2016a)]%
        {cityscapes_cvpr16}
\bibfield{author}{\bibinfo{person}{Marius Cordts}, \bibinfo{person}{Mohamed
  Omran}, \bibinfo{person}{Sebastian Ramos}, \bibinfo{person}{Timo Rehfeld},
  \bibinfo{person}{Markus Enzweiler}, \bibinfo{person}{Rodrigo Benenson},
  \bibinfo{person}{Uwe Franke}, \bibinfo{person}{Stefan Roth}, {and}
  \bibinfo{person}{Bernt Schiele}.} \bibinfo{year}{2016}\natexlab{a}.
\newblock \showarticletitle{The Cityscapes Dataset for Semantic Urban Scene
  Understanding}. In \bibinfo{booktitle}{\emph{Proc. of the IEEE Conference on
  Computer Vision and Pattern Recognition (CVPR)}}.
\newblock


\bibitem[Cordts et~al\mbox{.}(2016b)]%
        {cordts2016cityscapes}
\bibfield{author}{\bibinfo{person}{Marius Cordts}, \bibinfo{person}{Mohamed
  Omran}, \bibinfo{person}{Sebastian Ramos}, \bibinfo{person}{Timo Rehfeld},
  \bibinfo{person}{Markus Enzweiler}, \bibinfo{person}{Rodrigo Benenson},
  \bibinfo{person}{Uwe Franke}, \bibinfo{person}{Stefan Roth}, {and}
  \bibinfo{person}{Bernt Schiele}.} \bibinfo{year}{2016}\natexlab{b}.
\newblock \showarticletitle{The cityscapes dataset for semantic urban scene
  understanding}. In \bibinfo{booktitle}{\emph{Proceedings of the IEEE
  conference on computer vision and pattern recognition}}.
  \bibinfo{pages}{3213--3223}.
\newblock


\bibitem[Davis(2016)]%
        {davis2016theDS}
\bibfield{author}{\bibinfo{person}{Lennard~J. Davis}.}
  \bibinfo{year}{2016}\natexlab{}.
\newblock \bibinfo{booktitle}{\emph{The Disability Studies Reader (5th ed.)}}.
\newblock \bibinfo{publisher}{Routledge}.
\newblock
\urldef\tempurl%
\url{https://doi.org/10.4324/9781315680668}
\showDOI{\tempurl}


\bibitem[Deng et~al\mbox{.}(2009)]%
        {deng2009imagenet}
\bibfield{author}{\bibinfo{person}{Jia Deng}, \bibinfo{person}{Wei Dong},
  \bibinfo{person}{Richard Socher}, \bibinfo{person}{Li-Jia Li},
  \bibinfo{person}{Kai Li}, {and} \bibinfo{person}{Li Fei-Fei}.}
  \bibinfo{year}{2009}\natexlab{}.
\newblock \showarticletitle{Imagenet: A large-scale hierarchical image
  database}. In \bibinfo{booktitle}{\emph{2009 IEEE conference on computer
  vision and pattern recognition}}. Ieee, \bibinfo{pages}{248--255}.
\newblock


\bibitem[Everingham et~al\mbox{.}(2010)]%
        {everingham2010pascal}
\bibfield{author}{\bibinfo{person}{Mark Everingham}, \bibinfo{person}{Luc
  Van~Gool}, \bibinfo{person}{Christopher~KI Williams}, \bibinfo{person}{John
  Winn}, {and} \bibinfo{person}{Andrew Zisserman}.}
  \bibinfo{year}{2010}\natexlab{}.
\newblock \showarticletitle{The pascal visual object classes (voc) challenge}.
\newblock \bibinfo{journal}{\emph{International journal of computer vision}}
  \bibinfo{volume}{88} (\bibinfo{year}{2010}), \bibinfo{pages}{303--338}.
\newblock


\bibitem[Geiger et~al\mbox{.}(2012)]%
        {geiger2012we}
\bibfield{author}{\bibinfo{person}{Andreas Geiger}, \bibinfo{person}{Philip
  Lenz}, {and} \bibinfo{person}{Raquel Urtasun}.}
  \bibinfo{year}{2012}\natexlab{}.
\newblock \showarticletitle{Are we ready for autonomous driving? the kitti
  vision benchmark suite}. In \bibinfo{booktitle}{\emph{2012 IEEE conference on
  computer vision and pattern recognition}}. IEEE, \bibinfo{pages}{3354--3361}.
\newblock


\bibitem[Goyal et~al\mbox{.}(2017)]%
        {balanced_vqa_v2}
\bibfield{author}{\bibinfo{person}{Yash Goyal}, \bibinfo{person}{Tejas Khot},
  \bibinfo{person}{Douglas Summers{-}Stay}, \bibinfo{person}{Dhruv Batra},
  {and} \bibinfo{person}{Devi Parikh}.} \bibinfo{year}{2017}\natexlab{}.
\newblock \showarticletitle{Making the {V} in {VQA} Matter: Elevating the Role
  of Image Understanding in {V}isual {Q}uestion {A}nswering}. In
  \bibinfo{booktitle}{\emph{Conference on Computer Vision and Pattern
  Recognition (CVPR)}}.
\newblock


\bibitem[Gupta et~al\mbox{.}(2021)]%
        {Gupta2021GPV}
\bibfield{author}{\bibinfo{person}{Tanmay Gupta}, \bibinfo{person}{A. Kamath},
  \bibinfo{person}{Aniruddha Kembhavi}, {and} \bibinfo{person}{Derek Hoiem}.}
  \bibinfo{year}{2021}\natexlab{}.
\newblock \showarticletitle{Towards General Purpose Vision Systems}.
\newblock \bibinfo{journal}{\emph{ArXiv}}  \bibinfo{volume}{abs/2104.00743}
  (\bibinfo{year}{2021}).
\newblock


\bibitem[Gurari et~al\mbox{.}(2018)]%
        {gurari2018vizwiz}
\bibfield{author}{\bibinfo{person}{Danna Gurari}, \bibinfo{person}{Qing Li},
  \bibinfo{person}{Abigale~J Stangl}, \bibinfo{person}{Anhong Guo},
  \bibinfo{person}{Chi Lin}, \bibinfo{person}{Kristen Grauman},
  \bibinfo{person}{Jiebo Luo}, {and} \bibinfo{person}{Jeffrey~P Bigham}.}
  \bibinfo{year}{2018}\natexlab{}.
\newblock \showarticletitle{Vizwiz grand challenge: Answering visual questions
  from blind people}. In \bibinfo{booktitle}{\emph{Proceedings of the IEEE
  Conference on Computer Vision and Pattern Recognition}}.
  \bibinfo{pages}{3608--3617}.
\newblock


\bibitem[He et~al\mbox{.}(2017)]%
        {he2017mask}
\bibfield{author}{\bibinfo{person}{Kaiming He}, \bibinfo{person}{Georgia
  Gkioxari}, \bibinfo{person}{Piotr Doll{\'a}r}, {and} \bibinfo{person}{Ross
  Girshick}.} \bibinfo{year}{2017}\natexlab{}.
\newblock \showarticletitle{Mask {R-CNN}}. In
  \bibinfo{booktitle}{\emph{Proceedings of the IEEE international conference on
  computer vision (ICCV)}}.
\newblock


\bibitem[Islam et~al\mbox{.}(2024)]%
        {islam2024dataset}
\bibfield{author}{\bibinfo{person}{Md~Touhidul Islam}, \bibinfo{person}{Imran
  Kabir}, \bibinfo{person}{Elena~Ariel Pearce}, \bibinfo{person}{Md~Alimoor
  Reza}, {and} \bibinfo{person}{Syed~Masum Billah}.}
  \bibinfo{year}{2024}\natexlab{}.
\newblock \bibinfo{title}{A Dataset for Crucial Object Recognition in Blind and
  Low-Vision Individuals' Navigation}.
\newblock
\newblock
\showeprint[arxiv]{2407.16777}~[cs.CV]
\urldef\tempurl%
\url{https://arxiv.org/abs/2407.16777}
\showURL{%
\tempurl}


\bibitem[Li et~al\mbox{.}(2022a)]%
        {li2022lavis}
\bibfield{author}{\bibinfo{person}{Dongxu Li}, \bibinfo{person}{Junnan Li},
  \bibinfo{person}{Hung Le}, \bibinfo{person}{Guangsen Wang},
  \bibinfo{person}{Silvio Savarese}, {and} \bibinfo{person}{Steven~CH Hoi}.}
  \bibinfo{year}{2022}\natexlab{a}.
\newblock \showarticletitle{{Lavis}: A library for language-vision
  intelligence}.
\newblock \bibinfo{journal}{\emph{arXiv preprint arXiv:2209.09019}}
  (\bibinfo{year}{2022}).
\newblock


\bibitem[Li et~al\mbox{.}(2022b)]%
        {li2022blip}
\bibfield{author}{\bibinfo{person}{Junnan Li}, \bibinfo{person}{Dongxu Li},
  \bibinfo{person}{Caiming Xiong}, {and} \bibinfo{person}{Steven Hoi}.}
  \bibinfo{year}{2022}\natexlab{b}.
\newblock \showarticletitle{Blip: Bootstrapping language-image pre-training for
  unified vision-language understanding and generation}. In
  \bibinfo{booktitle}{\emph{International Conference on Machine Learning}}.
  PMLR, \bibinfo{pages}{12888--12900}.
\newblock


\bibitem[Li et~al\mbox{.}(2021)]%
        {li2021adversarial}
\bibfield{author}{\bibinfo{person}{Linjie Li}, \bibinfo{person}{Jie Lei},
  \bibinfo{person}{Zhe Gan}, {and} \bibinfo{person}{Jingjing Liu}.}
  \bibinfo{year}{2021}\natexlab{}.
\newblock \showarticletitle{Adversarial vqa: A new benchmark for evaluating the
  robustness of vqa models}. In \bibinfo{booktitle}{\emph{Proceedings of the
  IEEE/CVF International Conference on Computer Vision}}.
  \bibinfo{pages}{2042--2051}.
\newblock


\bibitem[Lin et~al\mbox{.}(2014)]%
        {lin2014microsoft}
\bibfield{author}{\bibinfo{person}{Tsung-Yi Lin}, \bibinfo{person}{Michael
  Maire}, \bibinfo{person}{Serge Belongie}, \bibinfo{person}{James Hays},
  \bibinfo{person}{Pietro Perona}, \bibinfo{person}{Deva Ramanan},
  \bibinfo{person}{Piotr Doll{\'a}r}, {and} \bibinfo{person}{C~Lawrence
  Zitnick}.} \bibinfo{year}{2014}\natexlab{}.
\newblock \showarticletitle{Microsoft coco: Common objects in context}. In
  \bibinfo{booktitle}{\emph{Computer Vision--ECCV 2014: 13th European
  Conference, Zurich, Switzerland, September 6-12, 2014, Proceedings, Part V
  13}}. Springer, \bibinfo{pages}{740--755}.
\newblock


\bibitem[Linton(1998)]%
        {linton1998claiming}
\bibfield{author}{\bibinfo{person}{Simi Linton}.}
  \bibinfo{year}{1998}\natexlab{}.
\newblock \bibinfo{booktitle}{\emph{Claiming disability: Knowledge and
  identity}}.
\newblock \bibinfo{publisher}{NYU Press}.
\newblock


\bibitem[Morris(2020)]%
        {morris2020ai}
\bibfield{author}{\bibinfo{person}{Meredith~Ringel Morris}.}
  \bibinfo{year}{2020}\natexlab{}.
\newblock \showarticletitle{AI and Accessibility}.
\newblock \bibinfo{journal}{\emph{Commun. ACM}} \bibinfo{volume}{63},
  \bibinfo{number}{6} (\bibinfo{year}{2020}), \bibinfo{pages}{35--37}.
\newblock


\bibitem[Neuhold et~al\mbox{.}(2017a)]%
        {mapiliary_iccv17}
\bibfield{author}{\bibinfo{person}{Gerhard Neuhold}, \bibinfo{person}{Tobias
  Ollmann}, \bibinfo{person}{Samuel~Rota Bul{\`o}}, {and}
  \bibinfo{person}{Peter Kontschieder}.} \bibinfo{year}{2017}\natexlab{a}.
\newblock \showarticletitle{The Mapillary Vistas Dataset for Semantic
  Understanding of Street Scenes}.
\newblock \bibinfo{journal}{\emph{2017 IEEE International Conference on
  Computer Vision (ICCV)}} (\bibinfo{year}{2017}), \bibinfo{pages}{5000--5009}.
\newblock


\bibitem[Neuhold et~al\mbox{.}(2017b)]%
        {neuhold2017mapillary}
\bibfield{author}{\bibinfo{person}{Gerhard Neuhold}, \bibinfo{person}{Tobias
  Ollmann}, \bibinfo{person}{Samuel Rota~Bulo}, {and} \bibinfo{person}{Peter
  Kontschieder}.} \bibinfo{year}{2017}\natexlab{b}.
\newblock \showarticletitle{The mapillary vistas dataset for semantic
  understanding of street scenes}. In \bibinfo{booktitle}{\emph{Proceedings of
  the IEEE international conference on computer vision}}.
  \bibinfo{pages}{4990--4999}.
\newblock


\bibitem[OKO(2023)]%
        {oko_app}
\bibfield{author}{\bibinfo{person}{OKO}.} \bibinfo{year}{2023}\natexlab{}.
\newblock \bibinfo{title}{OKO makes every intersection accessible}.
\newblock
\newblock
\urldef\tempurl%
\url{https://www.ayes.ai/oko}
\showURL{%
\tempurl}


\bibitem[Park et~al\mbox{.}(2021)]%
        {park2021designing}
\bibfield{author}{\bibinfo{person}{Joon~Sung Park}, \bibinfo{person}{Danielle
  Bragg}, \bibinfo{person}{Ece Kamar}, {and} \bibinfo{person}{Meredith~Ringel
  Morris}.} \bibinfo{year}{2021}\natexlab{}.
\newblock \showarticletitle{Designing an Online Infrastructure for Collecting
  AI Data From People With Disabilities}. In
  \bibinfo{booktitle}{\emph{Proceedings of the 2021 ACM Conference on Fairness,
  Accountability, and Transparency}}. \bibinfo{pages}{52--63}.
\newblock


\bibitem[Ren et~al\mbox{.}(2015)]%
        {ren2015faster}
\bibfield{author}{\bibinfo{person}{Shaoqing Ren}, \bibinfo{person}{Kaiming He},
  \bibinfo{person}{Ross Girshick}, {and} \bibinfo{person}{Jian Sun}.}
  \bibinfo{year}{2015}\natexlab{}.
\newblock \showarticletitle{Faster r-cnn: Towards real-time object detection
  with region proposal networks}.
\newblock \bibinfo{journal}{\emph{Advances in neural information processing
  systems}}  \bibinfo{volume}{28} (\bibinfo{year}{2015}).
\newblock


\bibitem[Richter et~al\mbox{.}(2017)]%
        {pfb_ICCV2017}
\bibfield{author}{\bibinfo{person}{Stephan~R. Richter},
  \bibinfo{person}{Zeeshan Hayder}, {and} \bibinfo{person}{Vladlen Koltun}.}
  \bibinfo{year}{2017}\natexlab{}.
\newblock \showarticletitle{Playing for Benchmarks}. In
  \bibinfo{booktitle}{\emph{{IEEE} International Conference on Computer Vision,
  {ICCV} 2017, Venice, Italy, October 22-29, 2017}}.
\newblock


\bibitem[Saha et~al\mbox{.}(2019)]%
        {saha2019project}
\bibfield{author}{\bibinfo{person}{Manaswi Saha}, \bibinfo{person}{Michael
  Saugstad}, \bibinfo{person}{Hanuma~Teja Maddali}, \bibinfo{person}{Aileen
  Zeng}, \bibinfo{person}{Ryan Holland}, \bibinfo{person}{Steven Bower},
  \bibinfo{person}{Aditya Dash}, \bibinfo{person}{Sage Chen},
  \bibinfo{person}{Anthony Li}, \bibinfo{person}{Kotaro Hara}, {et~al\mbox{.}}}
  \bibinfo{year}{2019}\natexlab{}.
\newblock \showarticletitle{Project sidewalk: A web-based crowdsourcing tool
  for collecting sidewalk accessibility data at scale}. In
  \bibinfo{booktitle}{\emph{Proceedings of the 2019 CHI Conference on Human
  Factors in Computing Systems}}. \bibinfo{pages}{1--14}.
\newblock


\bibitem[Seo and Jung(2017)]%
        {seo2017exploring}
\bibfield{author}{\bibinfo{person}{Woosuk Seo} {and} \bibinfo{person}{Hyunggu
  Jung}.} \bibinfo{year}{2017}\natexlab{}.
\newblock \showarticletitle{Exploring the community of blind or visually
  impaired people on YouTube}. In \bibinfo{booktitle}{\emph{Proceedings of the
  19th International ACM SIGACCESS Conference on Computers and Accessibility}}.
  \bibinfo{pages}{371--372}.
\newblock


\bibitem[Seo and Jung(2018)]%
        {seo2018understanding}
\bibfield{author}{\bibinfo{person}{Woosuk Seo} {and} \bibinfo{person}{Hyunggu
  Jung}.} \bibinfo{year}{2018}\natexlab{}.
\newblock \showarticletitle{Understanding blind or visually impaired people on
  youtube through qualitative analysis of videos}. In
  \bibinfo{booktitle}{\emph{Proceedings of the 2018 ACM International
  Conference on Interactive Experiences for TV and Online Video}}.
  \bibinfo{pages}{191--196}.
\newblock


\bibitem[Seo and Jung(2021)]%
        {seo2021understanding}
\bibfield{author}{\bibinfo{person}{Woosuk Seo} {and} \bibinfo{person}{Hyunggu
  Jung}.} \bibinfo{year}{2021}\natexlab{}.
\newblock \showarticletitle{Understanding the community of blind or visually
  impaired vloggers on YouTube}.
\newblock \bibinfo{journal}{\emph{Universal Access in the Information Society}}
   \bibinfo{volume}{20} (\bibinfo{year}{2021}), \bibinfo{pages}{31--44}.
\newblock


\bibitem[Simons and Levin(1997)]%
        {simons1997change}
\bibfield{author}{\bibinfo{person}{Daniel~J Simons} {and}
  \bibinfo{person}{Daniel~T Levin}.} \bibinfo{year}{1997}\natexlab{}.
\newblock \showarticletitle{Change blindness}.
\newblock \bibinfo{journal}{\emph{Trends in cognitive sciences}}
  \bibinfo{volume}{1}, \bibinfo{number}{7} (\bibinfo{year}{1997}),
  \bibinfo{pages}{261--267}.
\newblock


\bibitem[Theodorou et~al\mbox{.}(2021)]%
        {theodorou2021disability}
\bibfield{author}{\bibinfo{person}{Lida Theodorou}, \bibinfo{person}{Daniela
  Massiceti}, \bibinfo{person}{Luisa Zintgraf}, \bibinfo{person}{Simone
  Stumpf}, \bibinfo{person}{Cecily Morrison}, \bibinfo{person}{Edward Cutrell},
  \bibinfo{person}{Matthew~Tobias Harris}, {and} \bibinfo{person}{Katja
  Hofmann}.} \bibinfo{year}{2021}\natexlab{}.
\newblock \showarticletitle{Disability-first Dataset Creation: Lessons from
  Constructing a Dataset for Teachable Object Recognition with Blind and Low
  Vision Data Collectors}. In \bibinfo{booktitle}{\emph{The 23rd International
  ACM SIGACCESS Conference on Computers and Accessibility}}.
  \bibinfo{pages}{1--12}.
\newblock


\bibitem[Tigwell(2021)]%
        {tigwell2021nuanced}
\bibfield{author}{\bibinfo{person}{Garreth~W Tigwell}.}
  \bibinfo{year}{2021}\natexlab{}.
\newblock \showarticletitle{Nuanced perspectives toward disability simulations
  from digital designers, blind, low vision, and color blind people}. In
  \bibinfo{booktitle}{\emph{Proceedings of the 2021 CHI Conference on Human
  Factors in Computing Systems}}. \bibinfo{pages}{1--15}.
\newblock


\bibitem[Wang et~al\mbox{.}(2023)]%
        {wang2023yolov7}
\bibfield{author}{\bibinfo{person}{Chien-Yao Wang}, \bibinfo{person}{Alexey
  Bochkovskiy}, {and} \bibinfo{person}{Hong-Yuan~Mark Liao}.}
  \bibinfo{year}{2023}\natexlab{}.
\newblock \showarticletitle{YOLOv7: Trainable bag-of-freebies sets new
  state-of-the-art for real-time object detectors}. In
  \bibinfo{booktitle}{\emph{Proceedings of the IEEE/CVF Conference on Computer
  Vision and Pattern Recognition}}. \bibinfo{pages}{7464--7475}.
\newblock


\bibitem[Wang et~al\mbox{.}(2020)]%
        {wang2020deep2}
\bibfield{author}{\bibinfo{person}{Jingdong Wang}, \bibinfo{person}{Ke Sun},
  \bibinfo{person}{Tianheng Cheng}, \bibinfo{person}{Borui Jiang},
  \bibinfo{person}{Chaorui Deng}, \bibinfo{person}{Yang Zhao},
  \bibinfo{person}{Dong Liu}, \bibinfo{person}{Yadong Mu},
  \bibinfo{person}{Mingkui Tan}, \bibinfo{person}{Xinggang Wang},
  {et~al\mbox{.}}} \bibinfo{year}{2020}\natexlab{}.
\newblock \showarticletitle{Deep high-resolution representation learning for
  visual recognition}.
\newblock \bibinfo{journal}{\emph{IEEE transactions on pattern analysis and
  machine intelligence}} \bibinfo{volume}{43}, \bibinfo{number}{10}
  (\bibinfo{year}{2020}), \bibinfo{pages}{3349--3364}.
\newblock


\bibitem[Xie et~al\mbox{.}(2022)]%
        {xie2022youtube}
\bibfield{author}{\bibinfo{person}{Jingyi Xie}, \bibinfo{person}{Na Li},
  \bibinfo{person}{Sooyeon Lee}, {and} \bibinfo{person}{John~M Carroll}.}
  \bibinfo{year}{2022}\natexlab{}.
\newblock \showarticletitle{YouTube Videos as Data: Seeing Daily Challenges for
  People with Visual Impairments During COVID-19}. In
  \bibinfo{booktitle}{\emph{Proceedings of the 2022 ACM Conference on
  Information Technology for Social Good}}. \bibinfo{pages}{218--224}.
\newblock


\bibitem[Zhang et~al\mbox{.}(2023)]%
        {zhang2023recognize}
\bibfield{author}{\bibinfo{person}{Youcai Zhang}, \bibinfo{person}{Xinyu
  Huang}, \bibinfo{person}{Jinyu Ma}, \bibinfo{person}{Zhaoyang Li},
  \bibinfo{person}{Zhaochuan Luo}, \bibinfo{person}{Yanchun Xie},
  \bibinfo{person}{Yuzhuo Qin}, \bibinfo{person}{Tong Luo},
  \bibinfo{person}{Yaqian Li}, \bibinfo{person}{Shilong Liu}, {et~al\mbox{.}}}
  \bibinfo{year}{2023}\natexlab{}.
\newblock \showarticletitle{Recognize Anything: A Strong Image Tagging Model}.
\newblock \bibinfo{journal}{\emph{arXiv preprint arXiv:2306.03514}}
  (\bibinfo{year}{2023}).
\newblock


\bibitem[Zhou et~al\mbox{.}(2017)]%
        {zhou2017scene}
\bibfield{author}{\bibinfo{person}{Bolei Zhou}, \bibinfo{person}{Hang Zhao},
  \bibinfo{person}{Xavier Puig}, \bibinfo{person}{Sanja Fidler},
  \bibinfo{person}{Adela Barriuso}, {and} \bibinfo{person}{Antonio Torralba}.}
  \bibinfo{year}{2017}\natexlab{}.
\newblock \showarticletitle{Scene parsing through ade20k dataset}. In
  \bibinfo{booktitle}{\emph{Proceedings of the IEEE conference on computer
  vision and pattern recognition}}. \bibinfo{pages}{633--641}.
\newblock


\bibitem[Zhou et~al\mbox{.}(2019)]%
        {zhou2019semantic}
\bibfield{author}{\bibinfo{person}{Bolei Zhou}, \bibinfo{person}{Hang Zhao},
  \bibinfo{person}{Xavier Puig}, \bibinfo{person}{Tete Xiao},
  \bibinfo{person}{Sanja Fidler}, \bibinfo{person}{Adela Barriuso}, {and}
  \bibinfo{person}{Antonio Torralba}.} \bibinfo{year}{2019}\natexlab{}.
\newblock \showarticletitle{Semantic understanding of scenes through the ade20k
  dataset}.
\newblock \bibinfo{journal}{\emph{International Journal of Computer Vision}}
  \bibinfo{volume}{127} (\bibinfo{year}{2019}), \bibinfo{pages}{302--321}.
\newblock


\end{thebibliography}


\appendix
\section{Appendix: Video Analysis and Object Labeling Details}
\label{sec:analysis-and-labeling}

\begin{table*}[!t]
    \begin{center}
    \small{
        \begin{tabular}{l C{3.75cm}  C{1.0cm} C{0.50cm} C{1.0cm} C{0.75cm} C{1.25cm} C{3.75cm}}
            \toprule
             \textbf{ID} & 
             \textbf{Title/Context}  &
             {\textbf{Duration}} &
             {\textbf{\# Segments}} &
             {\textbf{\# Annotated Seg.}} &
             \textbf{Year} &
             \textbf{Location} &
             \textbf{URL} \\
             \toprule
            %
             V1 & Blind Man Walking & 2:24  & 5 & 2 & 2011 & London & \url{https://youtu.be/RmsoHyMRtbg}  \\ \hline
             \rowcolor{gray!10} 
             V2 & following a blind person for a day | JAYKEEOUT & 7:02 & 1& 1 &  2021 & Seoul & \url{https://youtu.be/dPisedvLKQQ} \\ \hline
             V3 & Orientation \& Mobility for the Blind-1* & 0:00-10:00  & 8 & 2 &  2012 & --- & \url{https://youtu.be/Gkf5tEbP-oo}\\ 
             \hline
             \rowcolor{gray!10} 
             V4 & Orientation \& Mobility for the Blind-2* & 10:01-19:10 & 4 & 3 & 2012 & --- & \url{https://youtu.be/Gkf5tEbP-oo?t=602} \\ 
             \hline
             V5 & My First Blind Cane Adventure to Get Coffee | Did I Succeed or Give Up* & 10:00 & 3 & 1 & 2019 & Caribbean Cruise Ship & \url{https://youtu.be/SZM-Le6MEE0} \\ 
             \hline              
             \rowcolor{gray!10} 
             V6 & Using A White Cane | Legally Blind* & 10:00 & 2 & 1 & 2018 & --- & \url{https://youtu.be/TxUxbXyh7Y4} \\ 
             \hline
             V7 & How a Blind Person Uses a Cane & 4:18 & 4 & 1 & 2013 & --- & \url{https://youtu.be/xi0JMS1rulo} \\ 
             \hline             
             \rowcolor{gray!10}              
             V8 & Orientation mobility & 9:36 & 2 & 1 & 2022 & --- & \url{https://youtu.be/6u53Q7IvVIY} \\ 
             \hline                           
             V9 & TAKING THE METRO AND WALKING THROUGH MADRID ALONE AND BLIND-1* & 9:19 & 4 & 1 & 2020 & Madrid & \url{https://youtu.be/Vx3-ltp9p-Y} \\ 
             \hline                                        
             \rowcolor{gray!10}  
             V10 & TAKING THE METRO AND WALKING THROUGH MADRID ALONE AND BLIND-2* & 10:00 - 19:00 & 1 & 1 & 2020 & Madrid & \url{https://youtu.be/Vx3-ltp9p-Y?t=600} \\ 
             \hline                           
             V11 & Mobility and Orientation Training for Young People with Vision Impairment & 5:48 & 3  & 1 & 2019 & Edinburgh & \url{https://youtu.be/u-3GlbJ5RMc} \\ 
             \hline                           
             \rowcolor{gray!10}  
             V12 & Mobility and Orientation & 8:49 & 4 & 1 & 2018 & New York City & \url{https://vimeo.com/296488214} \\ 
             \hline                           
             V13 & Traveling with the white cane & 2:14 & 3 & 1 & 2009 & Maryland & \url{https://vimeo.com/2851243} \\ 
             \hline              
             \rowcolor{gray!10} 
             V14 & Blindness Awareness Month - Orientation and Mobility with ELC and 1st Grade Students & 5:52 & 5 & 2 & 2022 & --- & \url{https://vimeo.com/758153786} \\ 
             \hline              
             V15 & The White Cane documentary & 5:40 & 3 & 1 & 2021 & --- & \url{https://vimeo.com/497359578} \\ 
             \hline              
             \rowcolor{gray!10} 
             V16 & Craig Eckhardt takes the subway on Vimeo	& 4:43 & 4 & 1 & 2010 & New York & \url{https://vimeo.com/17293270} \\ 
             \hline              
             V17 & Guide Techniques for people who are blind or visually impaired* & 10:00 & 3  & 2 & 2015 & --- & \url{https://youtu.be/iJfxkBOekvs} \\ 
             \hline              
             \rowcolor{gray!10} 
             V18 & Russia: Blind Commuter Faces Obstacles Every Day & 3:20 & 6 & 2 & 2013 & Moscow & \url{https://youtu.be/20W2ckx-BcE} \\ 
             \hline              
             V19 & The ``Challenges'' you may not know about ``Blind'' People | A Day in Bright Darkness & 8:00 & 6  & 2 & 2016 & Malaysia & \url{https://youtu.be/xdyj1Is5IFs} \\ 
             \hline              
             \rowcolor{gray!10} 
             V20 & Blind Challenges in a Sighted World & 3:54 & 5 & 2 & 2017 & --- & \url{https://youtu.be/3pRWq8ritc8} \\ 
             \hline              
             V21 & What to expect from Orientation \& Mobility Training (O\&M) at VisionCorps & 2:21 & 7 & 2 & 2012 & Pennsylvania & \url{https://youtu.be/wU7b8rwr2dM} \\ 
            \bottomrule
    \end{tabular}
    }
    \caption{List of our collected videos. We cropped the YouTube videos using \url{https://streamable.com}, which has a crop limit of 10 mins.} 
    \label{table:dataset}
    \end{center}
\end{table*}

\subsection{Video Analysis}
To thoroughly analyze the collected videos, 
we split each video into small clips of variable lengths between five and ninety-five seconds.
Each clip revolves around the appearance of objects that are significant in people's navigation on roads and sidewalks.
We refer to each clip as a video segment. 
Table~\ref{table:dataset} shows the number of segments created from each video. 
Using a keyframe extraction tool called \textit{Katna}\footnote{https://katna.readthedocs.io/en/latest/}, we further segmented the clips into keyframes. 
The keyframes are characterized as the representative frames of a video stream that serve a precise and concise summary of the video content, considering transitions in the scene and changes in lighting conditions and activities. 
The number of keyframes (i.e., images) extracted from the video segments was between three and ninety-three. 
Afterward, we manually annotated the presence and absence of objects (of our final list) for a subset of keyframes extracted from the video segments. 
Let us denote the final list of objects as $L_u$ for future reference.
Afterward, we manually labeled each video frame for each object in $L_u$.

\subsection{Object Labeling}
\label{subsec:object-labeling}
All authors of this paper visually inspected the keyframes to generate object labeling. 
Each author annotated a subset of video segments by observing the changes between two consecutive keyframes. 
For each frame, the existence (denoted by 1) or absence (denoted by 0) of all the objects that belong to $L_{u}$ was annotated. 
If In a keyframe $F_{k}$ the existence of the object $O \in L_{u}$ is denoted by a function $E$, the $E$ can be defined as follows:
\[
    E(O)= 
    \begin{cases}
        1,  & \text{if $O$ exists in $F_k$}\\
        0,  & \text{otherwise}
    \end{cases}
\]
Now the annotation of keyframe $F_k$ can be written as, $(E(O) \ | \ O \in L_u)$. 
Since visually inspecting changes can be subject to ``\textit{change blindness}''---a phenomenon when the visual feed is momentarily interrupted by a blank screen~\cite{simons1997change}---we avoided this by viewing two consecutive keyframe pairs on the screen side-by-side and glancing between them. 

For each video segment, annotating the first keyframe took the longest, typically 5-7 minutes. 
For subsequent frames, we only looked at what objects newly appeared/disappeared from the previous frame and adjusted the annotations accordingly.
Such differential annotation took less than 60 seconds in most cases. 
The annotation task required more time if a new frame appeared with a completely different background or camera viewport. 
With an average of around 15 keyframes per video segment, we needed around $\approx20$ minutes to label the objects a video. 
Each video segment was annotated by at least two researchers independently. 
Later, we resolved the conflicts in annotation through collaboration. 

\subsection{Analyzing Annotated Data}
\label{subsec:analyzing_data}
\begin{figure*}[!ht]
    \centering
    \includegraphics[width=1.0\linewidth]{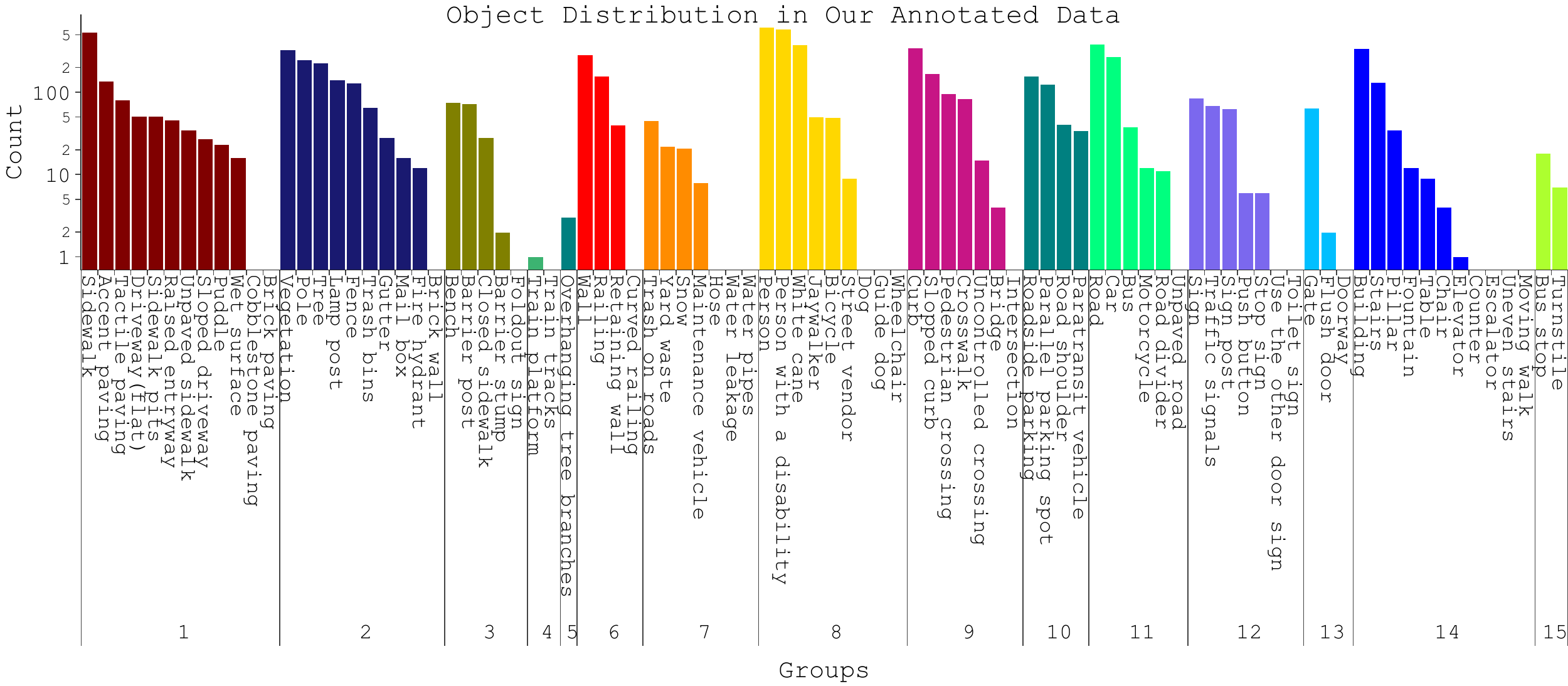}
    \caption{Bar chart representing the object distribution in our annotated data. Each bar represents the number of keyframes in which an object (as labeled on the x-axis) was present. 
    The X-axis also shows the id of the parent concept or group (as described in Table~\ref{table:taxonomy}) to which each object belongs. The Y-axis is in logarithmic scale.}
    \label{fig:dataset_stat}
\end{figure*}

The bar chart in Figure~\ref{fig:dataset_stat} depicts the distribution of objects in our annotated dataset.

\section{Appendix: Preliminary Evaluations Using Our Object Labeling}
\label{sec:assessing_ai_models}

\subsection{Model Selection}
We evaluated the preparedness of several state-of-the-art computer vision models for potential application in aiding blind and low-vision individuals utilizing our object labeling.
More specifically, we selected models specialized in addressing four distinct computer vision tasks: 
i) object recognition,
ii) object detection,
iii) semantic segmentation,
and iv) visual question answering (VQA). Table~\ref{table:model_type} shows the chosen models alongside their respective computer vision genre. 
We chose a total of seven models from these four types. 
For our experiments, all models were adapted to solve the recognition task.
The adaptation process is explained in Sec.~\ref{subsec:exp_procedure}.

\begin{table*}[!ht]
    \centering
\begin{tabular}{L{6cm} | C{6cm} } 
  \hline
   \textbf{\textsc{Type}} & 
   \textbf{\textsc{Models}}
   \\ 
  \hline
 \rowcolor{gray!10} 
  Recognition Model & RAM (Recognize Anything Model)~\cite{zhang2023recognize} \\
 \hline
  Detection Model & Faster R-CNN~\cite{ren2015faster}, YOLO V7~\cite{wang2023yolov7} \\
 \hline
 \rowcolor{gray!10} 
 Segmentation Model & HRNet V2~\cite{wang2020deep2}, Mask R-CNN~\cite{he2017mask}\\
 \hline
 VQA (Visual Question Answering)~\cite{antol2015vqa} Model & GPV-1~\cite{Gupta2021GPV}, BLIP~\cite{li2022blip, li2022lavis}\\
 \hline
\end{tabular}
    \caption{Selected models for evaluation and their type.}
    \label{table:model_type}
\end{table*}

While a recognition model only predicts the object categories present in a given image, detection models go beyond category prediction and additionally provide rectangular bounding boxes encompassing the objects.
On the other hand, semantic segmentation models detect fine-grained object masks (exact shapes of the objects) along with their categories. 
Traditionally, these models are trained on a set of pre-defined object categories and, hence, can not predict objects that were not present in the vocabulary set of training images.
More recently, the Recognize Anything Model (RAM) can predict open vocabulary categories due to its language decoder~\cite{zhang2023recognize}. 
Finally, VQA models consist of language encoders and decoders, enabling them to predict answers to open-ended questions. 
VQA models can also perform tasks such as image captioning, object localization, and image descriptions. 
Among all the models, RAM and VQA models are suitable for blind and low-vision individuals because of their open-ended capabilities. 

\subsection{Experiments and Observations}
\indent
\paragraph{Experiment Procedure.}
\label{subsec:exp_procedure}
Recognition models provide a list of predicted objects for a given image, which can be directly used for evaluation. However, other model types provide output in different formats requiring some pre-processing.

\textit{Recognition from Detection Model.} Traditionally, detection and segmentation models predict bounding boxes and masks, respectively, for each of the categories. If an object is present in a given image, then the detection model will provide us with the pixel coordinates of bounding boxes and the categories of detected objects. We ignored the bounding boxes in this experiment since we only needed the object categories. 

\textit{Recognition from Segmentation Model.} In contrast, the segmentation model predicts a mask for each object category. If an object exists in a frame, its corresponding mask will contain some nonzero values; otherwise, the whole mask will be a matrix of zeros. Thus, We easily created a list of objects that were present in a given image by analyzing the nonzero values of the masks.

\textit{Recognition from VQA Model.} For VQA models, we asked them a question for each of the objects regarding whether the object is present or not.
Generally, visual questions are formed from broad categories of question types such as (i) recognition, (ii) common sense, (iii) reasoning, (iv) OCR, and (v) counting. State-of-the-art VQA models perform well when asked visual recognition questions; however, these models are prone to failure when questions from other categories~\cite{li2021adversarial} are asked. To test our selected VQA models, we asked simple visual recognition questions. We generated a set of questions, each focusing on one accessibility-related object as shown in Table~\ref{table:taxonomy}. All of our questions followed a predominant structure like ``Is there object X in the scene?'', where X is an object that came from Table~\ref{table:taxonomy}. Consequently, the answer from the VQA model was either ``yes'' or ``no''. By analyzing the answers, we constructed the list of objects that were in the given image. 
We passed all the extracted keyframes (mentioned in Section~\ref{sec:video_collection}) to all the models we chose for this experiment and generated predictions following the steps mentioned earlier in this paragraph. 
Later, we matched the prediction of each model against our annotation.

\begin{table}[h]
\centering
\begin{tabular}{|c|c|c|c|c|}
\hline
Model & N & Precision & Recall & F1 \\
      &   & (Micro Avg.) & (Micro Avg.) & (Micro Avg.) \\
\hline
BLIP & 90 & 0.3366 & 0.8263 & 0.4783 \\
\rowcolor{gray!10}
GPV-1 & 90 & 0.2273 & 0.8070 & 0.3547 \\
RAM & 54 & 0.8175 & 0.2715 & 0.4077 \\
\rowcolor{gray!10}
YOLO V7 & 12 & 0.9437 & 0.1570 & 0.2692 \\
HRNet V2 & 15 & 0.6110 & 0.3998 & 0.4834 \\
\rowcolor{gray!10}
Mask R-CNN & 12 & 0.6077 & 0.1801 & 0.2779 \\
Faster R-CNN & 12 & 0.6029 & 0.1837 & 0.2815 \\
\hline
\end{tabular}
\caption{Precision, Recall, and F1 score (For all three metrics, higher is better) of all the selected models (shown in Table~\ref{table:model_type}) over our annotated keyframes. $N$ column shows the number of object categories the corresponding model can predict.}
\label{table:metric_summary}
\end{table}

\paragraph{Results and Observations.}
We calculated the micro average precision, recall, and F1 score of each of the chosen models. 
All the scores are reported in Table~\ref{table:metric_summary}. 
The table also shows the number of objects each model can predict under the $N$ column, out of $L_u$ objects in our dataset. 
From the table, we can observe that the detection and segmentation models can only predict labels for 12 to 15 objects from the list $L_u$---which explains their relatively poor performance.
The lack of representative annotations in their training dataset explains their inability to predict a larger number of object categories.

In contrast, BLIP, GPV-1, and RAM have a built-in language encoder/decoder. 
Therefore, they were able to predict more object categories. 
As we asked questions about the existence of each of the 90 object categories of list $L_u$ to BLIP and GPV-1, they generated answers for each question. 
On the other hand, RAM, being an open-ended object recognition model, could predict 54 object categories from our list. 
BLIP achieved the highest F1 score over 90 object categories among these three models.

\begin{figure*}[!ht]
    \centering
    \includegraphics[width=1.0\linewidth]{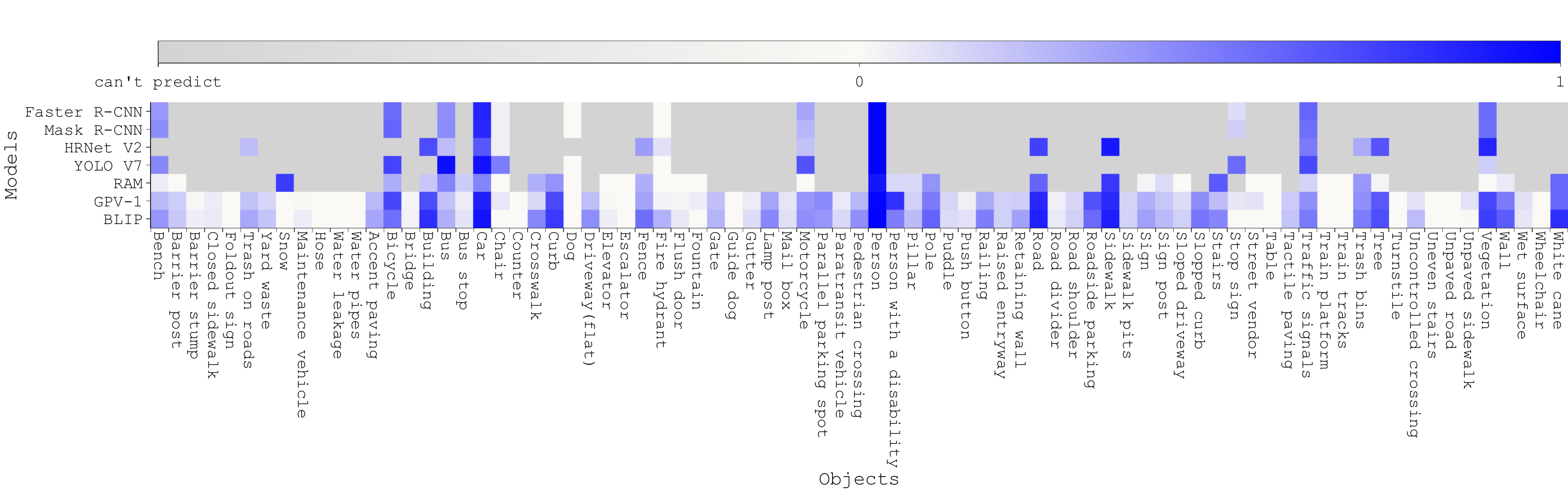}
    \caption{A heatmap representing the classwise F1 score of all the selected models (shown in Table~\ref{table:model_type}).}
    \label{fig:model_comp}
\end{figure*}

We also calculated classwise F1 scores of all the selected models and plotted a heatmap using these scores to visualize the classwise performance of all the models. Figure~\ref{fig:model_comp} shows the heatmap representing the classwise F1 score of all seven models. 
In this figure, \colorbox[HTML]{D3D3D3}{\phantom{x}} in a cell means the corresponding model can not predict the corresponding object.
The F1 score is within the $[0, 1]$ range and is represented by the intensity of the blue color ( \colorbox[HTML]{0000FF}{\phantom{x}} ) in Figure~\ref{fig:model_comp}. 
The heatmap shows that RAM, BLIP, and GPV-1 can recognize more objects than the detection and segmentation models. 
These models can accurately predict a subset of object categories listed in $L_u$, such as persons, cars, buildings, roads, sidewalks, etc. (object columns where the intensity of blue is higher). 

\begin{figure*}[!ht]
    \centering
    
    \includegraphics[width=0.5\linewidth]{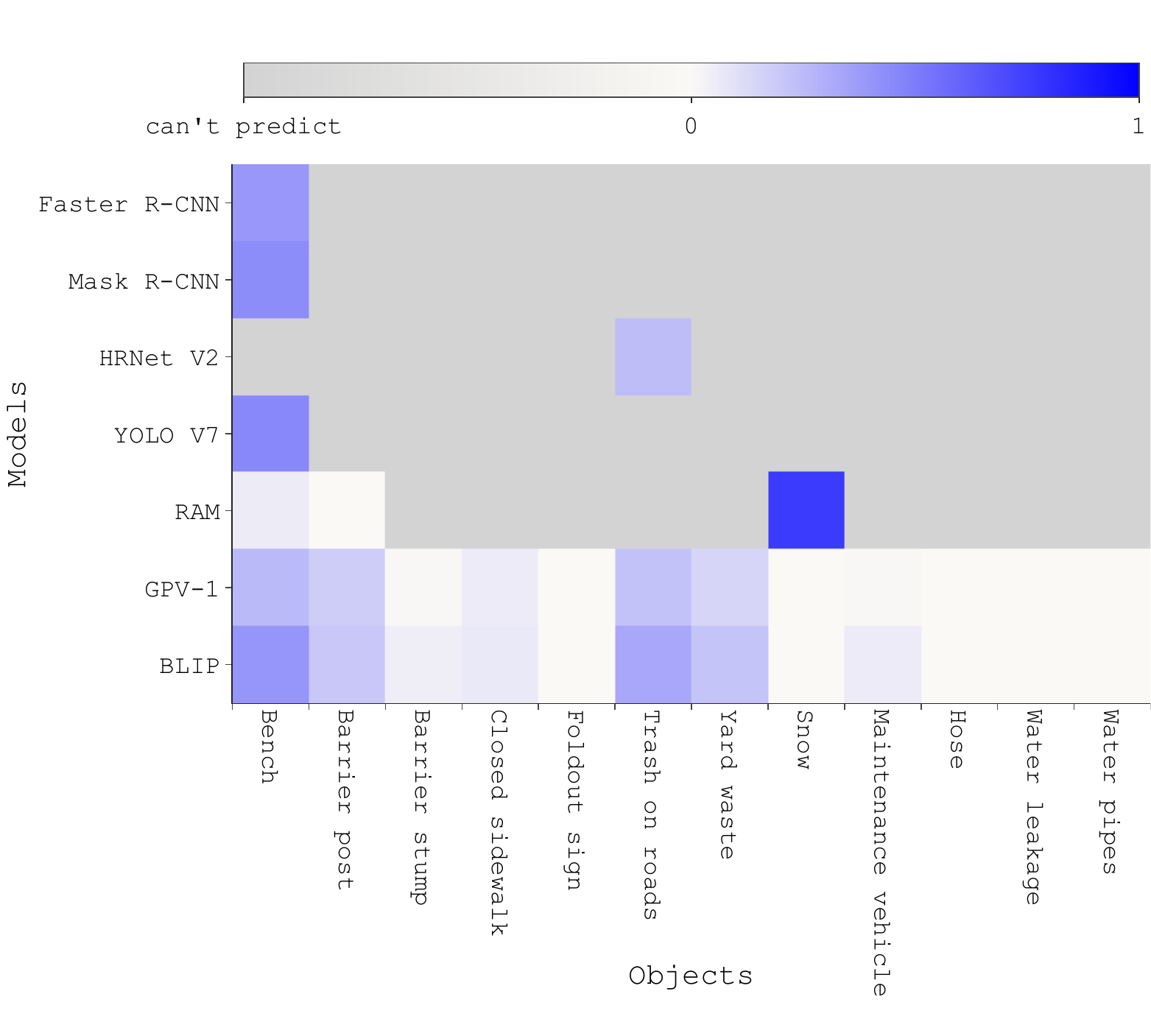}
    
    \caption{A heatmap representing the classwise F1 score of all the selected models for the objects of groups 3, 5, and 7 (shown in Table~\ref{table:taxonomy}).}
    \label{fig:model_comp_dang}
\end{figure*}

Moreover, none of the models can recognize the objects of the most significant groups (groups 3, 5, and 7) well. Figure~\ref{fig:model_comp_dang} represents the F1 scores of all the models for the objects of those three groups. 
The experiments described in this section suggest that any AI tool based on the selected models is not ready to be directly used in blind or low-vision individuals' navigation.

\renewcommand{\thefigure}{A\arabic{figure}}
\setcounter{figure}{0}

\renewcommand{\thetable}{A\arabic{table}}
\setcounter{table}{0}

\end{document}